\documentclass[
twocolumn,
hf,
]{ceurart}

\usepackage{listings}
\begin{document}

\copyrightyear{2023}
\copyrightclause{Copyright for this paper by its authors.
  Use permitted under Creative Commons License Attribution 4.0
  International (CC BY 4.0).}

\title{Solving ARC visual analogies with neural embeddings and vector arithmetic: A generalized method}

\author[1]{Luca H. Thoms}[%
orcid=0009-0001-7172-1337,
email=luca.thoms@student.uva.nl
]
\address[]{Psychological Methods, University of Amsterdam}

\author[1]{Karel A. Veldkamp}[%
orcid=0009-0009-7554-2262,
email=k.a.veldkamp@uva.nl
]

\author[1]{Hannes Rosenbusch}[%
orcid=0000-0002-4983-3615,
email=h.rosenbusch@uva.nl
]
\author[1]{Claire E. Stevenson}[%
orcid=0000-0003-1797-9836,
email=c.e.stevenson@uva.nl
]
\cormark[1]

\cortext[1]{Corresponding author}

\begin{abstract}
  Analogical reasoning derives information from known relations and generalizes this information to similar yet unfamiliar situations. One of the first generalized ways in which deep learning models were able to solve verbal analogies was through vector arithmetic of word embeddings, essentially relating words that were mapped to a vector space (e.g., king – man + woman =\_\_?). In comparison, most attempts to solve visual analogies are still predominantly task-specific and less generalizable. This project focuses on visual analogical reasoning and applies the initial generalized mechanism used to solve verbal analogies to the visual realm. Taking the Abstraction and Reasoning Corpus (ARC) as an example to investigate visual analogy solving, we use a variational autoencoder (VAE) to transform ARC items into low-dimensional latent vectors, analogous to the word embeddings used in the verbal approaches. Through simple vector arithmetic, underlying rules of ARC items are discovered and used to solve them. Results indicate that the approach works well on simple items with fewer dimensions (i.e., few colors used, uniform shapes), similar input-to-output examples, and high reconstruction accuracy on the VAE. Predictions on more complex items showed stronger deviations from expected outputs, although, predictions still often approximated parts of the item's rule set. Error patterns indicated that the model works as intended. On the official ARC paradigm, the model achieved a score of 2\% (cf. current world record is 21 \%) and on ConceptARC it scored 8.8\%. Although the methodology proposed involves basic dimensionality reduction techniques and standard vector arithmetic, this approach demonstrates promising outcomes on ARC and can easily be generalized to other abstract visual reasoning tasks.
\end{abstract}


\maketitle

\section{Introduction}
Analogies help clarify the relationship between two distinct entities by juxtaposing them to a similar paring whose relationship is well known. In other words, an analogy can explain how two entities relate by generalizing the rules of a better-known and comparable relationship. Importantly, analogical reasoning lies at the heart of higher-order cognition, often used in contexts such as argumentation or clarification \cite{Gentner18}. A straightforward measure to assess analogical reasoning in humans and machines is the a:b::c:d task, a paradigm requiring agents to align a with c to infer d through b \cite{Goswami91, Mikolov13a}. For example, the analogy “\textbf{man} is to \textbf{king} as \textbf{woman} is to \_\_? (queen)” requires us to think about the relationship between the concepts, realizing that gender is the deviating factor. Although humans have little problem relating different entities to form analogies, artificial intelligence (AI) systems struggle to form such generalizations that go beyond their training data \cite{Mitchell21}. 

\begin{figure}
  \centering
  \includegraphics[width=0.88\linewidth]{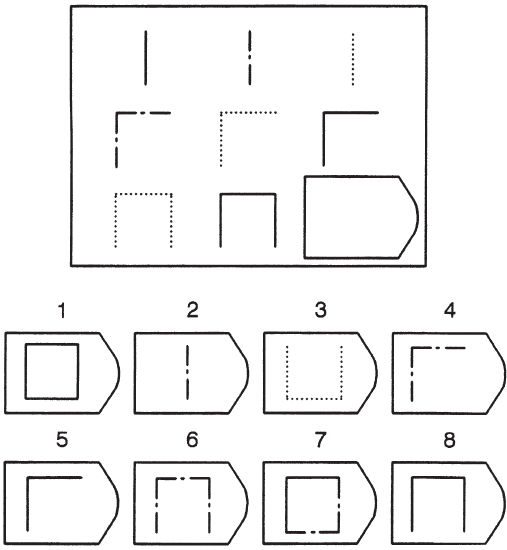}
  \caption{Raven’s Progressive Matrices, a non-verbal test, requires Agents to identify the relationship between visual features of sub-images and choose an answer that completes the matrix. The shown problem has the solution: No. 6 \cite{Raven03}.}
  \label{fig:raven}
\end{figure}

This project focuses on the domain of visual analogical reasoning, often assessed with abstract visual reasoning (AVR) tasks (e.g., Fig.~\ref{fig:raven}) such as those used in both human and AI tests of general intelligence. These tasks usually require agents to identify patterns between images containing two-dimensional shapes varying in their visual attributes (e.g., color, position). Through advancements in neural network architectures, a multitude of models can now solve complex AVR tasks, such as Raven’s Progressive Matrices or Bongard Problems \cite{Malkinski23}. However, this recent success is largely due to an inflation in training data available to train these models or hard-coded solutions, not their ability to generalize beyond what is known \cite{Malkinski22}. We need a generalized approach to solving AVR tasks. Interestingly, the first stepping stones enabling large language models to solve verbal analogies involved vector arithmetic, offering a generalizable method to tackle analogical reasoning in AI \cite{Mikolov13a}. How can we extend this approach to the visual realm?

\subsection{The Abstraction and Reasoning Corpus}
\begin{figure}
  \centering
  \includegraphics[width=0.9\linewidth]{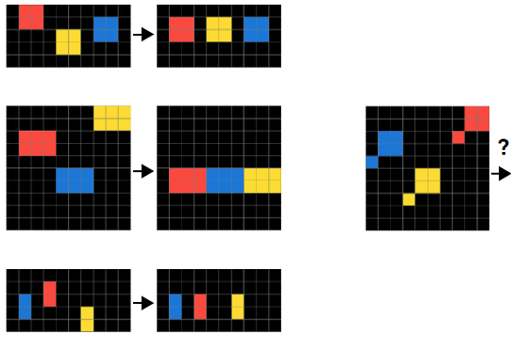}
  \caption{Each ARC item has a set of input-to-output examples and an unsolved input. The illustrated item has three example cases, showcasing the rule: The red and yellow shapes must be aligned vertically with the blue shape. (\url{https://arc-visualizations.github.io/index.html}).}
  \label{fig:arc_example}
\end{figure}

First, we need a task capable of testing model abilities in terms of visual analogical reasoning. This task should allow us to recreate the complexity of analogical processes while offering conditions to adapt modern machine learning procedures. The Abstraction and Reasoning Corpus (ARC) is a benchmark data set best suited for that purpose \cite{Chollet19}. The ARC provides 800 publicly available and unique items, which provides a wide variety of challenges. Unlike most AVR tasks that concern classification problems or multiple-choice items, ARC problems are open-ended, requiring the agent to create the response themself. Importantly, each item showcases a unique set of rules, resulting in a near-unlimited space of potential item rules and making it especially difficult to construct task-specific hard-coded solvers. These aspects limit a model's ability to exploit shortcuts in the data to arrive at solutions in a non-human way and assume generalization capabilities. As every item only provides a limited number of examples displaying its rule, solving them requires an established knowledge base using concepts like geometry, quantification, and relations \cite{Malkinski23, Odouard22}. As the example in figure~\ref{fig:arc_example} demonstrates, agents have to create the output from scratch, specifying size and color dimensions, compared to simply choosing a solution like in figure~\ref{fig:raven}. By enforcing few-shot learning, broad generalization in test data, and a generative output, ARC provides an ideal benchmark to investigate higher-order cognitive processes, such as visual analogy solving \cite{Johnson21}.

\subsection{Current approaches to solving ARC}
Thus far, attempts at solving ARC have focused primarily on symbolic approaches (i.e., human-readable representation of a task), with probabilistic program induction being the most common one \cite{Alford21, Fischer20, Xu22}. In this framework, a hand-crafted domain-specific language specifies a task's grammar (e.g., “move\_down”, “split\_colors”). Depending on the task, a probability distribution with respective priors is defined over this grammar space. Guided by this distribution, the grammar is sampled to find a solution for the task. However, this approach depends on defining the task-specific actions manually, in the form of logic statements \cite{Mitchell21}. This makes the models task-specific and difficult to generalize, for example, to other AVR domains. For reference, the current world record on ARC solves 21\% of the data set with a hand-crafted symbolic approach and over 7k C++ lines of code \cite{Quarks20}. When tested on slight variations of ARC items it was able to solve, it failed to fully generalize \cite{Odouard22}.

To date, \cite{Webb20} seem to be the only study to present a generalized method to solve multiple AVR tasks, mimicking symbolic approaches through a connectionist framework, i.e. using artificial neural networks without pre-programmed rules. They developed a neural network that processes components of simple AVR items (i.e., separate parts of an image) sequentially and binds similarities in external memory. Their algorithm was able to solve four AVR tasks using less training data than other approaches. However, the tested tasks concerned classification problems that could be processed sequentially, which is not applicable to ARC. Currently, no published approach has been able to solve ARC to a satisfying extent using artificial neural networks without pre-defined rules.

\subsection{Current study}
We investigate whether it is possible to solve ARC with a connectionist perspective. Word embedding methods from the field of natural language processing (e.g., word2vec) showcase the ability of language models to implicitly learn vector-space word representations \cite{Mikolov13b}. Surprisingly, these representations effectively capture semantic structures within language. This enabled earlier language models to solve verbal analogies with vector operations that rely solely on the differences between words. For example, the vector operation “King – Man + Woman” approximates the word vector “Queen” \cite{Mikolov13a}. Inspired by these past advancements, we explore whether an analogous visual approach can be used to solve ARC. Similar attempts on simpler visual analogies yielded promising results \cite{Reed15}. This approach could simplify ARC's complex visual analogy solving as it would involve simple vector operations on the generated embeddings (i.e., latent vector, representation) and can serve as a general foundation for solving other AVR tasks.

There are two challenges when using this approach. First, we have to create visual embeddings. We propose to encode ARC grids into neural embeddings using a convolutional encoder $f(x)$. Like in “word2vec”, the encoder maps inputs to a vector space \cite{Mikolov13b}. Given a vocabulary of related grids (${a\rightarrow b}$, ${c\rightarrow d}$), the vector $f(b)$ – $f(a)$ would capture the underlying rule, which we simply add to the query vector $f(c)$ in the embedding space. A decoder structure $g(x)$ maps the resulting embedding vector back to the original grid dimensions: 

\begin{displaymath}
    d = g(f(b) - f(a) + f(c)).
\end{displaymath}

Ideally, this vector transformation and subsequent mapping approximate the desired grid structure $d$ \cite{Reed15}. We use an autoencoder architecture to train the encoder and decoder simultaneously on ARC example grids (input and output) to accurately compress grid embeddings.

The second challenge is combining the multiple input-to-output examples per ARC item into one rule vector ($f(b)$ – $f(a)$). We investigate two approaches: The first approach is to consider all examples and take the average of all rule vectors. The second approach is to use the rule vector of examples with input most similar to the test input. In this case, the Euclidean distance is used to indicate closeness between input representations, exploiting the fact that grids are embedded in a latent vector space. To our knowledge, this is the first attempt to combine latent vector representations of visual analogy elements with vector arithmetic to solve an AVR task as challenging as ARC. Our main research question is:

\begin{enumerate}[1.]
\item Can we (partly) solve novel ARC items by creating separate latent vector representations for ARC input and output patterns, and exploit the differences in these representations to capture the underlying transformation process?
\end{enumerate}

\noindent
Secondly, given the number of ways to combine multiple input-to-output examples per item:

\begin{enumerate}[2.]
\item What combination of rule vectors (i.e., average-based versus similarity-based) yields the best solutions?
\end{enumerate}

\noindent
Lastly, ARC items differ in their complexity which might influence results:

\begin{enumerate}[3.]
\item What item characteristics (e.g., grid size changes, used colors across examples) drive the accuracy of the proposed approach?
\end{enumerate}

\section{Methodology}
\subsection{Data}
The ARC items (N = 800) can be read in and processed as matrices using Python. Each item provides two to ten example input grids matched with their respective example output grids. Grids can be any size from 1x1 to 30x30, and each cell is filled with one of ten possible colors. Agents infer from the small set of input-to-output examples the underlying rule that transforms the input into the output and apply this rule to a novel input grid to solve the item \cite{Chollet19}.

The official ARC data contains 400 training and 400 evaluation items. Subsequent training procedures for the autoencoder used a random subset of the official training data for training purposes (N = 300) and the withheld subset for validating the model (N = 100). As every item consists of multiple example input and output grids, more grids are at our disposal to train and validate the autoencoder. For example, if an item has three input-to-output examples (e.g., Fig.~\ref{fig:arc_example}), this provides us with six grids. Therefore, the actual number of data points available for building the autoencoder is considerably higher, amounting to around $\sim$ 1.950 grids for training and $\sim$ 650 grids for validation purposes. However, for the sake of clarity, we will refer to the general number of ARC items in future sections and not their specific grid compositions. We reserved the official evaluation items (N = 400) for later testing of the final ARC-solver.

\subsection{Creating embeddings with a variational autoencoder}
We use an autoencoder to create the grid embeddings. Autoencoders are a special type of generative neural network typically used for dimensionality reduction, mapping inputs into a vector space. The most basic form is composed of an encoder and a decoder. Both parts have different objectives, however, are trained and adjusted simultaneously. The encoder network learns to transform the original high-dimensional input into a low-dimensional latent vector embedding. This latent vector can be imagined as a flattened numerical vector, ideally having fewer dimensions than the input while still capturing the original information, comparable to a zip version of a file. By limiting the encoding to a smaller latent space, the model often learns salient features of the data since it must decide which parts of the input to keep, effectively encoding image semantics. Therefore, latent vectors of similar inputs should be similar, thus, closer in the vector space. This process is learned in reverse by the decoder network: It tries to reconstruct the original high-dimensional input from the learned latent representation, analogous to unzipping a zipped file. In essence, the autoencoder learns to minimize the differences between input and reconstruction \cite{Goodfellow16, Skansi18}.

In an autoencoder, the generation of output is purely deterministic and only constrained by the dimensions of the latent space. To expand the latent space beyond what is already known, we add a probabilistic component to the autoencoder, creating a variational autoencoder (VAE). We achieve this by projecting the input first onto a probability distribution and sampling the latent vector from this distribution. The variability in the sampled latent vectors allows the model to retrieve different latent vectors every time and increases the space of attainable solutions. An added benefit is the additional constraint to our loss function, the Kullback-Leibler divergence, enabling further regularization towards more general patterns within the data \cite{Girin21}.

\begin{figure*}
  \centering
  \includegraphics[width=\linewidth]{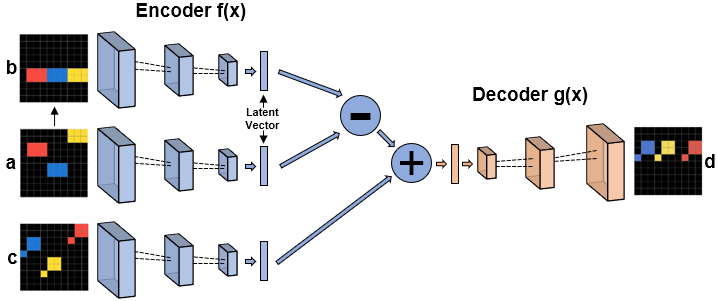}
  \caption{After creating latent vectors for each grid, we subtract the input representation $f(a)$ from the output representation $f(b)$. We add the result to the novel input representation $f(c)$ to solve it and decode it back into the original dimensions.}
  \label{fig:vas}
\end{figure*}

Once the VAE creates informative latent representations of item grids, separate latent vectors can be computed for an item’s example inputs and outputs, and combined similarly to word vectors: The item's input representations are subtracted from the item's output representations. This results in multiple rule vectors per item, describing its rule numerically. According to the chosen approach, the rule vectors are combined into one. We can add this vector to the unsolved input representation to create the corresponding output representation. From there, the trained VAE decoder generates the output grid prediction. Figure~\ref{fig:vas} displays the full procedure executed by the visual analogy solver (VAS).

\subsection{Training of the variational autoencoder}
\begin{figure}[b]
  \centering
  \includegraphics[width=\linewidth]{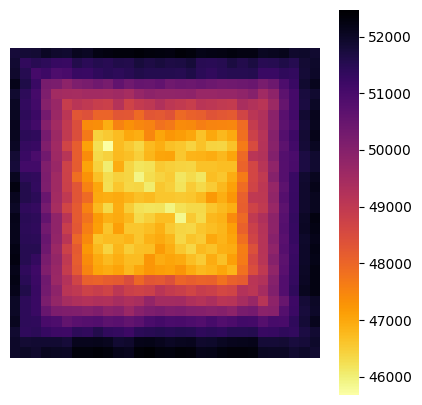}
  \caption{The heatmap illustrates the accuracy of individual pixel assignments by overlaying how many times they were reconstructed with the correct color on all evaluated grids (scaled to 30x30). A trend shows that pixels near edges were reconstructed more often correctly, indicating the distribution of shapes within items and the effects of zero-padding.}
  \label{fig:heat}
\end{figure}

For the current project, a VAE is constructed, with its full architectural details outlined in the following paragraphs. The model construction involved a combination of task-specific pre-processing steps, data augmentations, hyperparameter tuning, and validation on unseen training data. Training progress was monitored using auxiliary tools such as reconstruction accuracies across item grids and heatmaps of correct pixel assignments (e.g., Fig.~\ref{fig:heat}), ensuring the network learns appropriately and creates accurate reconstructions for each grid.

\paragraph{Data pre-processing}
Conventional convolutional neural networks (CNN) have difficulties processing inputs of varying sizes. Therefore, we scaled all ARC grids to the maximum item size of 30x30 but retained the proportions of the original grids by computing the Kronecker product over each pattern \cite{Weisstein}. If the transformation did not yield the desired dimensions, the grids were zero-padded (i.e., placing zeros at the edges to achieve 30x30). This allowed us to consider all items uniformly on the largest grid size. Lastly, grid colors were one-hot encoded (i.e., dummy coding), extending the final dimensions of each grid to 10 (colors) x 30 (height) x 30 (width). The initializing layer of the autoencoder considered the increased dimensionality by handling ten input channels. This allowed the network to process colors explicitly.

\paragraph{Data post-processing}
Consequently, the VAS (Fig.~\ref{fig:vas}) only processes and outputs grids of size 10x30x30 and does not fully consider size as a dimensionality. Hence, predictions made by the VAS are rescaled to the correct output grid size using the expected output grids.

\paragraph{Data augmentation}
As the data available to train the model (N = 300) is not enough for highly parameterized deep learning models, three data augmentations were implemented, helping the autoencoder learn more accurate representations of each grid. The first augmentation created copies of each grid pair with different colors. The second augmentation rotated 60\% of the items by 90, 180, or 270 degrees. Lastly, we created mirrored copies of each grid by reflecting its left-hand side at the midpoint. Core features of ARC were retained, providing additional examples for robust training of the latent dimensions. When applied together, the augmentations increased the training data from around $\sim$ 1.950 to $\sim$ 40.000 grids.

\paragraph{Hyperparameter tuning}
During the hyperparameter tuning process, we experimented with different combinations of layer counts, layer sizes, kernel sizes, and strides. Next to the mentioned augmentation techniques, we explored regularization strategies such as L2 regularization, denoising techniques (i.e., augmentation introducing noise to pictures), and beta-regularized versions to improve the network's ability to generalize to new data \cite{Goodfellow16, Higgins17}. Ultimately, a model with three (transposed) convolutional layers for the encoder and decoder, respectively, each containing 128 filters, yielded the best reconstructions. A kernel size of four and stride of two resulted in a reduction of the original image dimensions of 10x30x30 to 128x2x2 after passing through the encoder. 

Reducing the resolution from 30x30 to 2x2 improved generalization in the network compared to less drastic reductions (e.g., 30x30 to 22x22), likely reducing overfitting. Although regularization techniques enable models to generalize more efficiently beyond training data, in our case “unseen data” might refer to ARC items with new rules, shapes, and alignments. This introduces unknown dimensionality, making common autoencoder regularization methods less useful when evaluated on unseen items, mostly resulting in no noticeable improvements or more inaccurate reconstructions. However, a slight L2 regularization (penalty = 0.2) led to improvements when evaluated on the withheld data. 

Lastly, the number of latent dimensions requires careful consideration given our outlined approach. Although a higher latent vector space yields better reconstructions as it offers more dimensions to represent each grid, high-dimensional vector spaces are challenging, particularly when using vector arithmetic. Essentially, each grid representation is a point in this vector space after being encoded. The more dimensions we introduce, the farther apart these representations end up, making distance measures, a vital part of our proposal, less reliable as most data points will be far apart. This problem is often referred to as the curse of dimensionality, most noticeable in approaches like K-nearest neighbors utilizing vector spaces \cite{James13}. Tests demonstrated that 128 latent dimensions produced accurate reconstructions while maintaining relatively low dimensionality.

\paragraph{Model validation}
The autoencoder’s performance was validated on a separate, withheld subset of the official training data (N = 100), resulting in $\sim$ 650 grids for validation purposes. Auxillary tools such as reconstruction accuracies and heatmaps (Fig.~\ref{fig:heat}) assisted in the process. Given the above configurations, the autoencoder accurately reconstructed essential features of the data. It accurately reconstructed large, homogeneous areas of colors but struggled to reconstruct more nuanced colored shapes (e.g., narrow colored lines), resulting in a loss of information on some items. 

\begin{table*}
  \caption{Cell Prediction Accuracy across Data Sets and Rule Vector Approaches}
  \label{tab:acc}
  \begin{tabular}{ccccc}
    \toprule
    Grid Size & \multicolumn{2}{c}{Average Rule Vector} & \multicolumn{2}{c}{Similarity Rule Vector} \\
    \cmidrule(lr){2-3}
    \cmidrule(lr){4-5}
    & Training & Evaluation & Training & Evaluation\\
    \midrule
    Predicted 30x30 & 69.3\% & 70.84\% & 67.55\% & 69.27\% \\
    Predicted Rescaled & 61.19\% & 60.5\% & 60.09\% & 58.65\% \\
    Zero Filtered 30x30 & 37.17\% & 33.25\% & 36.89\% & 32.91\% \\
    Zero Filtered Rescaled & 37.01\% & 33.27\% & 37.13\% & 32.8\% \\
    \bottomrule
\end{tabular}
\end{table*}

\section{Results}
We tested the VAS on the full ARC data, including the training (N = 400) and evaluation (N = 400) set. Both include novel test inputs not seen by the model. However, example input and output grids corresponding to the novel test inputs of the training data were used to train and validate the autoencoder. This testing regime allowed us to investigate both partially seen and new rule objects. We assessed the prediction accuracy for test items as the share of correctly colored cells of the predicted grid compared to the expected test outputs. Final accuracy scores (see Tab.~\ref{tab:acc}) represent the average over items and are computed for four grid size conditions across data sets and rule vector approaches. Next to the predicted 30x30 grids (row \#1), we also investigated accuracy metrics on rescaled grids with the expected grid size (row \#2). However, certain task characteristics introduce a bias in this metric: One of the most common colors is black (i.e., cells containing zero) as it is frequently used as the grid background. Certain items are mostly black and have a small proportion of colored cells. Moreover, occasional padding when scaling to 30x30 introduced additional black cells to the grid. Consequently, some of the higher item accuracy scores are driven by large “empty” grids for which the model accurately predicts zeros but incorrectly assigns the few colored spots. To address this issue, we also report accuracy scores excluding black cells (rows \#3 and \#4), resulting in a drop in scores as evident in Table~\ref{tab:acc}. Generally, we see higher accuracy scores when using average rule vectors, a trend noticeable across data sets and grid size conditions. 

\begin{figure*}
  \centering
  \includegraphics[width=0.84\linewidth,trim=4 4 4 4,clip]{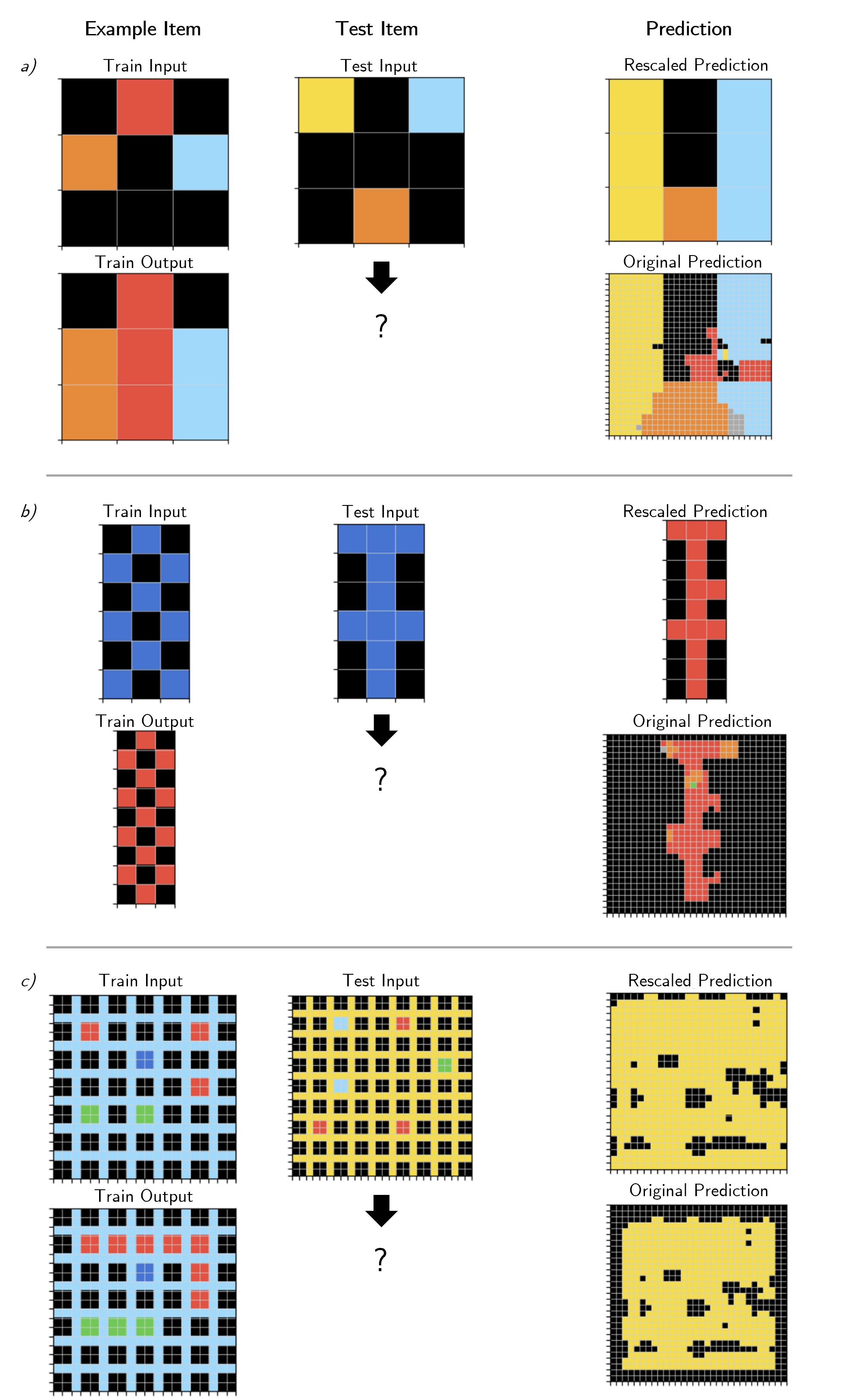}
  \caption{Showcasing the limitations of the VAS in predicting output grids across ARC items of differing complexity. Each item displays an example input-to-output, the corresponding unsolved test input, and the solver’s original and rescaled predictions. Items range from simple (a) over moderate (b) to advanced (c) difficulty levels, successively increasing the rule’s complexity. Predictions are based on the average rule vector approach.}
  \label{fig:spot}
\end{figure*}

\subsection{Item spotlight}
To provide an intuition of how the solver works, we will showcase its performance on individual items using the average rule vector approach. Notably, the model performed well on items with few rules and low complexity. Figure~\ref{fig:spot}a shows an example of this, featuring an item that requires agents to fill in everything below a colored cell with this very color. All input-to-output examples display this simple rule over a 3x3 grid. Although the model’s original prediction contains some cells it did not correctly predict, the rescaled version corresponds to the required solution, effectively denoising the grid. However, with an increased item complexity, the solver’s predictions get less accurate. Figure~\ref{fig:spot}b illustrates an item whose rule involves multiple components, such as color changes, grid expansions, and differing shapes between train and test items. While this might be an easier item for humans, the model struggles to expand the shape accurately, yet captures the color change and expansion well, showing that it reproduces parts of the item's rule. Nevertheless, the more detailed the grids and complex the rules get, the more the solver struggles as shown in the last example (see Fig.~\ref{fig:spot}c). Not only does this rule require the connection of same-colored squares, but the “lines” drawn between objects involve only the empty squares between matching colors. On top, the grid is further divided into a narrow-colored, yellow grid. For humans this might again be straightforward, however, the model struggles as the predictions show. The many involved dimensions combined with a challenging reconstruction make it difficult for the solver to go beyond generalizing the presence of a yellow grid.

The above items are just illustrative. Generally, the VAS captured the rule sets on certain items well enough to approximate solutions. When allowing variation within the model (i.e., sampling of latent vectors from the probability distribution during encoding), it can fully solve (involving rescaling) 20 and 14 items from the training and evaluation set, respectively. Although the model's predictions were farther from the expected solutions on high-complexity items, they sometimes captured components relevant to the rule in question (see Fig.~\ref{fig:spot}c). This suggests that the orientation captured by an item’s rule vectors gave a reasonable approximation of the general location of solutions within the vector space, albeit not very accurate. As proof of concept, consider items requiring the generation of a 1x1 output grid filled with only one color. To understand how the model arrives at a solution, imagine the grid representations as individual data points within a vector space: First, we subtract the input from the output grid representation, resulting in a transformation vector (i.e., direction within the vector space) describing their relationship. The model adds this (rule) vector to the unsolved input grid representation (i.e., another single point in the vector space) and ends up in an area of the vector space describing the wanted color.

\subsection{Performance on Chollet’s official ARC paradigm}
\cite{Chollet19} allows test-takers a maximum of three attempts per item on the official ARC test paradigm. The final score is determined by the fraction of items the test-taker successfully solves within the evaluation set, hence, rules the agent has no prior knowledge of. Importantly, the restriction on the evaluation set stems from the idea that algorithms might learn from the training data whereas humans would most likely not need this training phase to solve evaluation items. This allows the algorithms to make themselves familiar with the task dimensions \cite{Chollet19}. To adapt our algorithm to this challenge, we allowed the model to sample the latent vectors after the encoding process from a probability distribution, resulting in slightly different encodings every trial. This provided the solver with more variation, albeit random, instead of fixing its predictions to the mean of the distribution for every attempt as done for the initial accuracy scores (Tab.~\ref{tab:acc}). In other words, the attempts vary because of the probabilistic encoding. Moreover, we rescaled predictions and considered the rescaled grids as final predictions. Finally, we predicted the first test input of items since most have only one. Given these configurations, the VAS correctly predicted 8 out of the 400 official evaluation items (2\%) on three attempts. Both rule vector approaches amounted to the same score, however, deviated slightly in the items solved. This score is considerably lower than measured human performance (M = 83.8\%) \cite{Johnson21} and the score achieved by the current best-performing model on ARC (21\%), a hand-crafted symbolic algorithm reaching the first place at the original ARC Kaggle competition \cite{Quarks20}. However, the human participants only attempted 10 out of 40 randomly selected items belonging to the training set, deviating from the intended paradigm \cite{Johnson21}. 

Recently, \cite{Moskvichev23} published ConceptARC, a data set with newly constructed ARC items (N = 160) that are supposed to evaluate more human-like concept abstraction. They divided the tasks into 16 core concepts each with ten items \cite{Moskvichev23}. Further, items have three different test inputs, however, our predictions will only consider the first one to represent the item. Following the previous conditions, the VAS correctly predicted 14 out of the 160 ConceptARC items (8.8\%) on three attempts. The average rule vector approach solved more items compared to similarity-based predictions. Table~\ref{tab:concept} in the appendix features the accuracy per concept group of the two rule vector approaches in comparison to human performance from \cite{Moskvichev23}. Here, accuracy refers to the correctly solved fraction of all test inputs in the corresponding concept group (N = 30).

\subsection{Errors made by the VAS}
An interesting aspect of ARC is the types of errors produced by test-takers. As ARC requires agents to generate the output from scratch, the mistakes made are informative about the strategies used and understanding of rules. Recent studies that qualitatively assessed human errors on ARC showed that participants’ erroneous solutions were often relatively close to the expected outputs, featuring the most relevant components of rules, such as shapes, colors, or alignments \cite{Johnson21}. By examining mistakes made by the algorithm, we want to emphasize how the solver works and draw parallels to the representations humans use. We will look at how the two rule vector approaches differ in this regard, allowing variation in the sampled latent vectors to capture a range of errors. We focus on two items \cite{Johnson21} exposed their participants to and showcase example errors made by the participants. Interested readers may consult the highlighted study to also examine errors produced by the aforementioned symbolic algorithm on the same items.

\begin{figure*}
  \centering
  \includegraphics[width=\linewidth]{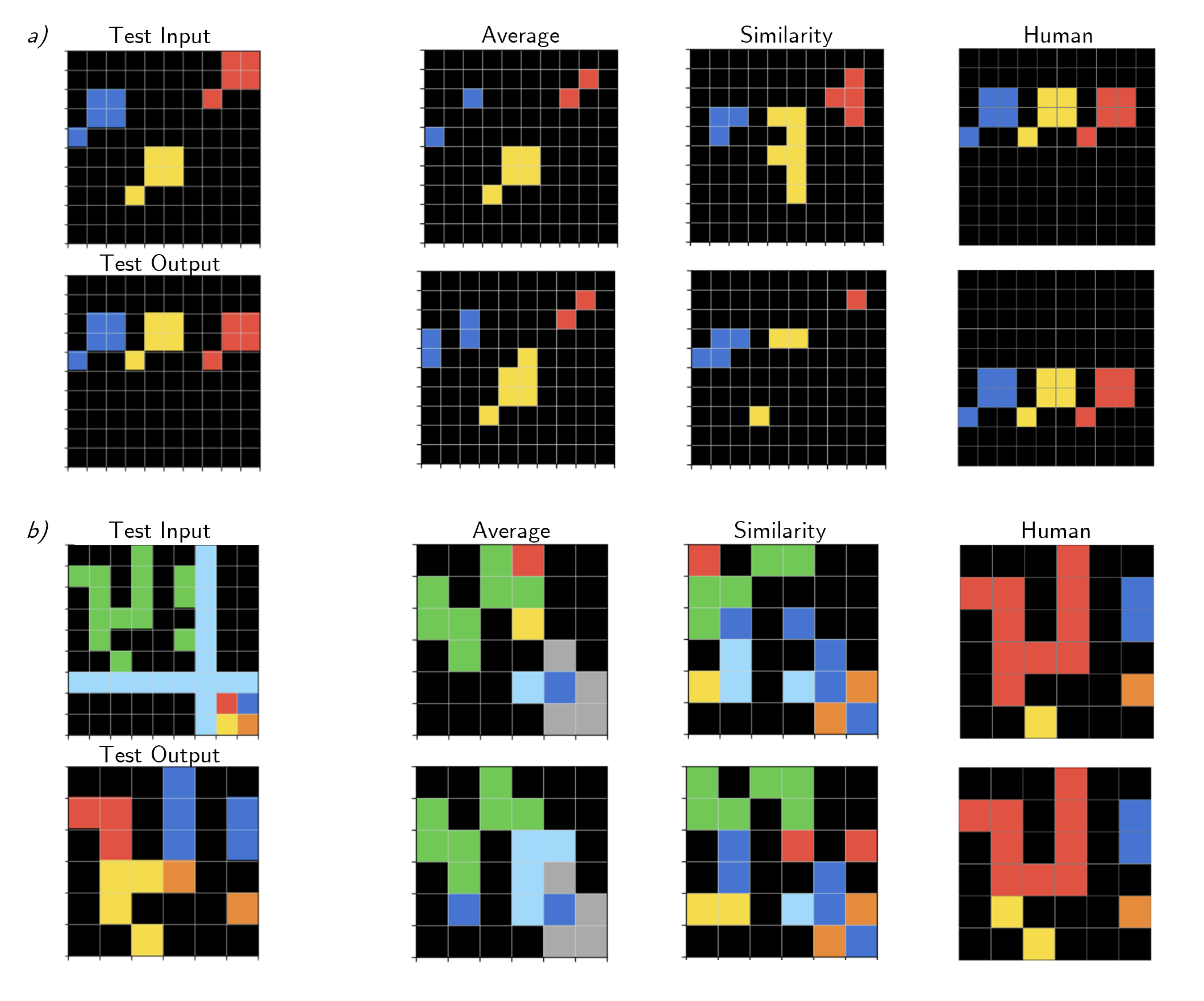}
  \caption{Errors made by the VAS and humans on two items of differing complexity. For each item, we display the provided test input grid and expected test output grid as well as two example predictions made by each rule vector approach and a human. (\url{https://arc-visualizations.github.io/index.html}).}
  \label{fig:error}
\end{figure*}

The first item is considered low in complexity, asking test-takers to simply align structures vertically with the blue shape (see Fig.~\ref{fig:error}a). Evidently, none of the rule vector techniques fully solved this task. Still, both methods inferred the correct color dimensions and sometimes even the distinctive shapes and alignment procedures, with the similarity approach approximating the solution best. Both similarity-based predictions attempt to align the yellow shapes vertically with the blue box as indicated by the repositioned, yellow-colored cells. Hence, we can assume that the rule vector derived from the few input-to-output examples captures this shift in colors, informing the model that blue cells are not moved in the process. Nevertheless, all predictions violate priors set by the input grid regarding objects’ shape, displaying elongated and cut-off versions.

The second item is considered more complex, requiring agents to apply the color scheme from the separated corner to the pattern in the corresponding separation (see Fig.~\ref{fig:error}b). Clearly, the solver did not find the correct solution. A common trend throughout all predictions is the accumulation of green cells in the top right corner and the center positioning of light blue cells, regardless of the rule vector method. It seems that the algorithm did not move the initial input representation far from its original position in the vector space. This is particularly evident when examining the average rule vector approach. One explanation could be that the example rule vectors were highly dissimilar, thus, pointing in opposite directions. This may have resulted in an average vector with little directional information. Yet, the similarity-based approach demonstrates better generalization patterns of rule dimensions, positioning colors in the correct quadrants as intended by the apply-color-pattern rule.

Both examples support the notions of earlier findings: Less complex items are better approximated than more difficult ones. The proposed vector arithmetic captures certain elements of the grids and input-to-output transformation, allowing the model to weakly adhere to rules such as color shifts and object positioning. In contrast, humans seem to make mistakes on these items too, but their errors are less severe, and solutions still resemble the most relevant features for given rules (see Fig.~\ref{fig:error}). Although the solver displays some generalization capabilities regarding rules, errors are comparatively more egregious, a fact that could be traced back to factors such as the encoding quality of grids or the simplicity of the proposed vector arithmetic.

\subsection{Factors driving accuracy}
To assess the influence of item and procedure characteristics on our model’s accuracy, we conducted an ordinary least squares regression with multiple predictors. As the model was partially trained and validated on the training data, the following analysis will focus on the evaluation data. First, we formulated several features that numerically describe the complexity of an item. This includes the number of provided input-to-output examples per item, if there is a grid size change between an item’s example inputs, the average grid size and the average number of colors used within an item’s inputs and outputs, the average rate of change and the average percent of zeros (i.e., color black) displayed within an item’s input and output grids, and the average similarity between an item’s inputs and outputs as measured by matching cells. Further, we binary coded whether there is a grid size and color change between an item’s inputs and outputs as well as between an item’s training inputs and test input. Next, we generated features relevant to the procedure that potentially influenced the model’s performance. This involved the average scaling factor per item input and output and accounted for how much we scaled individual grids to the desired grid size of 30x30. Lastly, we computed the average VAE reconstruction accuracy on each item, assessing how well the VAE encodes an item. By regressing accuracy on these variables, we can identify which factors contributed to the solver’s performance.

\paragraph{Feature selection}
To identify the most relevant features for explaining accuracy, we applied several feature selection techniques, including the inspection of standardized coefficient estimates, forward stepwise regression, cross-validated recursive feature elimination, and a LASSO regularized model fit. Due to the number of features, the feature selection techniques had strict selection criteria. Consequently, we used the full LASSO penalty (weight = 1.0), a threshold of 0.01 (p-value) for stepwise regressions, and 300 folds for recursive feature elimination to reduce selected features to the most relevant ones. We fitted regressions on accuracy measures concerning the 30x30 and rescaled prediction grids (see Tab.~\ref{tab:acc}) across different rule vectors, resulting in four separate regressions. As an example, table~\ref{tab:reg} shows the regression of the accuracy concerning the 30x30 grids using the average rule vector approach. The remaining three regression tables can be found in the corresponding code notebook (\textit{Visual Analogy Solving - ARC}) of the first author's GitHub repository (\url{https://github.com/foger3/ARC_DeepLearning}).

\begin{table*}
  \caption{Regression: 30x30 Grid Accuracy using the Average Rule Vector}
  \label{tab:reg}
  \begin{tabular}{cccccc}
    \toprule
    Feature & Estimate (std.) & \textit{SE} & \multicolumn{2}{c}{95\% CI} & \textit{p} \\
    \cmidrule(lr){4-5}
    & & & \textit{Lower} & \textit{Upper} &\\
    \midrule
    Average\_Similarity & 0.6806 & 0.053 & 0.576 & 0.785 & 0.000 \\
    Average\_Reconstruction & 0.4807 & 0.077 & 0.330 & 0.632 & 0.000 \\
    Average\_Scale\_Y & -0.1456 & 0.040 & -0.224 & -0.067 & 0.000 \\
    Average\_RoC\_X & 0.1988 & 0.102 & -0.001 & 0.399 & 0.051 \\
    Average\_Size\_X & -0.2048 & 0.107 & -0.415 & 0.005 & 0.056 \\
    Average\_Zeros\_Y & 0.0863 & 0.049 & -0.009 & 0.182 & 0.076 \\
    Average\_Zeros\_X & -0.0749 & 0.045 & -0.164 & 0.014 & 0.099 \\
    Grid\_Size\_Change & 0.0732 & 0.046 & -0.017 & 0.163 & 0.110 \\
    Color\_Change\_T & -0.0473 & 0.030 & -0.106 & 0.012 & 0.115  \\
    Average\_RoC\_Y & -0.1447 & 0.095 & -0.331 & 0.042 & 0.128 \\
    Average\_Size\_Y & 0.1322 & 0.110 & -0.084 & 0.348 & 0.229 \\
    Average\_Colors\_X & -0.0458 & 0.053 & -0.150 & 0.058 & 0.388 \\
    Average\_Scale\_X & -0.0243 & 0.043 & -0.110 & 0.061 & 0.575 \\
    Color\_Change & -0.0164 & 0.031 & -0.076 & 0.044 & 0.592 \\
    Average\_Colors\_Y & -0.0226 & 0.050 & -0.122 & 0.077 & 0.655 \\
    Grid\_Size\_Change\_T & -0.0189 & 0.047 & -0.112 & 0.074 & 0.690 \\
    Size\_Differences & -0.0032 & 0.048 & -0.098 & 0.092 & 0.947 \\
    Number\_Examples & -0.0019 & 0.031 & -0.063 & 0.059 & 0.951 \\
    \bottomrule
\end{tabular}
\end{table*}

\paragraph{Across models}
Unanimously, across models and selection methods, the features average similarity, average reconstruction accuracy, and average percent of zeros within an item's example grids were consistently ranked as the most relevant features for explaining accuracy. Based on the coefficients of the first feature (\textit{Average\_Similarity}), we see that the more similar the input and output examples are, the higher the accuracy of predictions. Intuitively, this makes sense as highly similar examples indicate easier transformations between inputs to outputs, thus, a less complex rule. If the average reconstruction accuracy of an item is high (\textit{Average\_Reconstruction}), then the model's accuracy on this item is also high. This suggests that the encoding quality of grids plays an important role in solving items. Being able to accurately reconstruct items is a vital part of our proposal as it entails appropriate grid representations with the latent vector space. Consequently, if representations loosely match their original inputs, the solver cannot accurately infer the transformation from inputs to outputs. This may result in predictions ending up at positions in vector space distant from the intended solution. 

\begin{figure}[b]
  \centering
  \includegraphics[width=\linewidth]{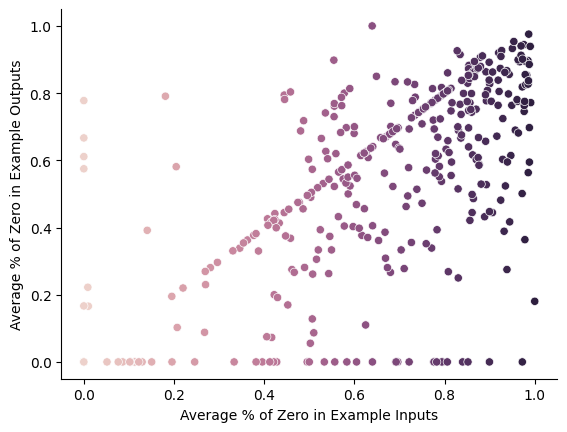}
  \caption{The figure displays the average percent of zeros (i.e., color black) in example inputs versus example outputs of the evaluation set. Noticeable is the lower right triangle, including most of the data points. This trend points to the observation of fewer zeros in example outputs with more sparse example inputs.}
  \label{fig:scatter}
\end{figure}

We observed opposing effects when investigating the coefficients of the average percent of zeros feature for example inputs (\textit{Average\_Zeros\_X}) versus example outputs (\textit{Average\_Zeros\_Y}) in table~\ref{tab:reg}. Logically, the more zeros appear in example outputs, the more zeros we would expect in the test output. Consequently, accuracy scores of predictions would go up as we saw earlier the accuracy metric is inflated due to the many zeros used as background in expected outputs. The solver might inaccurately label the few colored cells but correctly identifies many expected zeros, resulting in a high accuracy score. This effect is noticeable when looking at the coefficient of the average percent of zeros in example outputs as it has a positive direction. However, we observe a negative direction for the average percent of zeros in example inputs, indicating that more zeros in example inputs lead to lower accuracy scores on predictions. Keeping in mind the design of ARC items, we see that when example inputs are sparse (i.e., have a lot of black/zeros) the corresponding outputs often do not have as much black. For example, an item might ask the test-taker to zoom into a certain colorful structure or fill in empty objects on a sparse input grid, resulting in an output with less black. The lower right triangle of the scatterplot in Figure~\ref{fig:scatter} supports this notion. A general observation of predictions shows that the solver struggles to infer rules on these types of items due to reconstruction difficulties of the mainly black but more nuanced grids and its general tendency to predict colored cells less accurately than black cells.

\paragraph{Across grid sizes}
Comparing the selected features between regressions on 30x30 and rescaled accuracy scores showed an interesting trend involving item scaling and size factors. Regressions concerning the accuracy of the predicted 30x30 grids highlighted the scaling factors of items (\textit{Average\_Scale\_X}, \textit{Average\_Scale\_Y}). The higher the scaling factor, the lower the accuracy on a given item. Resizing a small item grid results in colored cells being increased from individual cells to a collection of cells. Therefore, in cases where the expected output grid dimensions are small, our model predicts the output on an enlarged scale, increasing the likelihood of predicting noise in the cluster of cells reflecting the single cell. This finding supports our idea of effectively denoising predictions by rescaling the 30x30 grid to the expected grid size. Upon examining the regressions of the rescaled accuracy scores, feature selection indicated that the occurrence of size differences between an item’s example (\textit{Size\_Differences}) inputs is relevant. As we withheld information regarding the size dimension by scaling and padding input and output grids independently, the resulting transformation vectors did not capture size differences between examples. This resulted in the model not being able to consider grid size differences between item examples, subsequently, decreasing the model's performance on rescaled grids. Task feature selection highlights the trade-off we made between the size dimension of ARC and modern CNN architecture's limitation of only processing same-sized inputs. 

\paragraph{Across rule vectors}
Lastly, we compared regressions between the two rule vector approaches. A consistent difference regarding the selected features concerned the average number of colors used within an item’s inputs (\textit{Average\_Colors\_X}) and outputs (\textit{Average\_Colors\_Y}). Especially for explaining accuracy scores based on similarity rule vectors, this feature was considered relevant. Precisely, the more colors an item uses per grid, the lower the resulting accuracy score, independent of scaling. As the similarity approach chooses the rule vector belonging to the example input most similar to the test input, the technique chooses a specific transformation. More colors indicate more dimensionality and potential variations between items. In this case, even the most similar input and their corresponding transformations might be very different from the presented test input. Consequently, this effect might be more pronounced for similarity-based accuracy as average-driven ones can compensate for this discrepancy. Nonetheless, this trend is noticeable across both approaches.

\section{Discussion}
The findings suggest that a VAE-based algorithm free of hard-coded rules performs well on a certain subset of visual analogy tests. We observed that items with simple rule vectors and low complexity, such as one-colored solutions, were closely approximated by the model. Although the model struggled with complex rules and high dimensional grids, it often captured important item characteristics (e.g., colored grids) or rule components (e.g., object repositioning) (Fig.~\ref{fig:spot}). Additionally, the errors it generated were often closely related to the task at hand (Fig.~\ref{fig:error}). Investigating both predictions and errors illustrates an important aspect of the model: The solver (VAS) does not predict grid cells in a chaotic way or independently from each other. Because the model treats test inputs as starting positions within the mapped vector space ($f(c)$), it creates a new starting point for every prediction. By adding a transformation term ($f(b)$ – $f(a)$) derived from a few examples to this new origin, the final prediction is always associated with the original baseline. These transformation vectors are adjusted for every item, adding another level of specificity for individual predictions (i.e., few-shot learning). Importantly, the quality of input and output mappings to the low dimensional vector space largely depends on a separate network (VAE) that can be fine-tuned to the task at hand. Thus, the model tackles the challenges imposed by ARC, such as broad generalizations and few-shot learning, by splitting them across model components: While the VAE captures general task dimensions, the VAS allows for the processing of the few item examples. Above all, we maintained a purely connectionist approach and demonstrated that a vector operation as simple as the subtraction of vectors can capture complex relationships and approximate item solutions of a task (ARC) considered very challenging for modern algorithms. Beyond the ARC-specific implications, we provided a general framework for solving analogies, extendable to other AVR tasks.

Another important question concerned the performance of the two rule vector approaches. Results from the task feature analyses and visual inspections of item predictions indicated that similarity-based performance suffered more from the increased complexity of items. If more dimensions are involved and item examples differ, even the most similar example input might be considerably different from the corresponding test input, consequently, requiring a different transformation. Although average-based rule vectors might compensate, their performance is not much better in those cases. Interestingly, the error analysis showed a noticeable trend in favor of similarity rule vectors: Especially on easier items the similarity-based approach seemed to capture rules better than its counterpart. Due to a low dimensional context, the chosen rule vector might be a good approximation of the true transformation while offering a definite direction (see Fig.~\ref{fig:error}a). In the case of average rule vectors, however, the averaging of vectors might result in a more restrained transformation with little directional information as we observed in Figure~\ref{fig:error}b. Generally, judging by accuracy scores (Tab.~\ref{tab:acc}), both approaches seem to perform comparably. Yet, looking at individual predictions the similarity rule vector approach seems to provide a slightly better generalization of rule characteristics. 

\subsection{Limitations and future directions}
While the simple combination of embeddings through vector arithmetic presented here showed promising results, the below list of extensions might prove powerful in future endeavors.

\paragraph{Simplicity of the procedure}
First, it can be argued whether the model is capable of analogical reasoning and equipped to reproduce a cognitive task this complex. Combining vectors through simple arithmetic operations might be too simple to mimic brain processes this intricate. In the present research, we only considered a simple addition of unsolved input representations and rule vectors. A common extension of this simple approach is to replace the simple vector arithmetic with a network model that can be trained in a supervised way. By training another separate network on the intended merge of rule vectors and input representations, hidden combinatorial patterns could be addressed, something the current model conveniently ignores \cite{Reed15}. Additionally, the presented rule vector approaches were chosen for convenience as they logically adapted our vector arithmetic methodology to task-specific problems such as the multiple input-to-output examples provided by each ARC item. Certainly, combinations of the multiple rule vectors could consider factors such as item complexity, a metric we may approximate using the cosine similarity between rule vectors, or item-specific rule vector combinations based on specific item characteristics, like a weighted average rule vector based on the similarity between example inputs. Exploring how to merge rule vector strategies goes beyond the general versions we presented. But, it does seem necessary to address inter-item differences and the general complexity of ARC items to improve our approach.

\paragraph{Data processing}
Furthermore, certain data pre-processing steps require improvements in the future. As mentioned, the model does not consider size differences between grids due to scaling all patterns to the same size. Not only does this violate the intended ARC paradigm, but findings indicate that the model performs worse on items with size differences between example grids. Although the scaling retains positional information vital for convolutional layers and avoids extensive padding of grids, every grid is expanded by its own scaling factor, sometimes resulting in differing factors between inputs and outputs. Future extensions should test whether the same scaling factor between example inputs and outputs leads to more sensible item solutions and retains parts of the size dimension. 

\paragraph{Model configurations}
Adding to this, a future adaptation of the model could incorporate these scaling factors in its computations to address size differences between examples and the respective input-to-outputs. For example, scaling could be included as a weighting factor in the proposed merging of rule vectors and their combination with input representations through an MLP. 

Another important feature driving model performance concerned the VAE’s reconstruction accuracy. The current encoding seems inadequate in capturing the full range of task dimensions. Given the simple, custom VAE architecture we used, pre-trained encoder and decoder networks might capture the underlying dimensions better, allowing more accurate representations of grids with nuanced shapes. Moreover, incorporating state-of-the-art structures such as spatial transformers into the convolutional architecture could improve embeddings by adding an attentive mechanism \cite{Jaderberg15}. Multiple domains come to mind when thinking about potential expansions of the proposed methodology. 

\paragraph{Generalizability}
Lastly, we only tested the proposed approach on one out of many AVR tasks. Even though ARC is considered one of the more difficult benchmarks in the field of visual analogical reasoning, a validation of our model on other, easier AVR tasks is missing. Certain components of the model would require adjustments as the original ARC images were encoded in a specific data format (i.e., JSON). However, this only concerns task-specific pre-processing and vector combinations, not the proposed core methodology. Accordingly, future research should expand our efforts to different AVR tasks.

\paragraph{Performance}
The visual analogy solver offers a simple approximation to a complex problem. Evidently, its performance lacks when compared to human reasoning. Although only a subset of the items was considered, prior research on ARC items demonstrated that humans outperform purely connectionist approaches to such problems. This is noticeable when we consider the impact of similarity between example inputs and outputs on the accuracy of the solver. Even on items with major differences between input and output, humans achieve interpretable and often correct predictions \cite{Johnson21}. In contrast, the solver still does not reach correct predictions on items considered easy due to their highly similar inputs and outputs, although it may be close. Its predictions on complex items deviate even more from expected outputs, a trait that is not as pronounced in human solutions (Fig.~\ref{fig:error}). Moreover, our final predictions were based on a rescaled version of the predicted 30x30 grids. When testing the solver on inputs with unknown outputs, we cannot rescale to the expected grid size. Rescaling here merely showed a denoised and potential prediction, given the model would adjust its output based on size differences.

\section{Conclusion}
The presented methodology is straightforward, involving simple dimensionality reduction techniques and ordinary vector arithmetic, and shows promising results on a challenging task like the ARC. Due to its simplicity and generalizability in conveying visual meaning numerically, it can be easily adapted to other AVR tasks, allowing future efforts to test the framework on new tasks, items, and dimensions. Successful implementation of this approach on other AVR tasks would provide research on visual analogies with a generalized framework and highlight the necessity for similar approaches in the field. Importantly, the approach does not incorporate any hard-coded rules and provides a fully connectionist solution (\textit{ARC-Zero}) with minimal implementation effort. Lastly, the present research successfully adapted a technique from the field of natural language processing, showing that even cross-domain challenges involving both verbal and visual components could follow the proposed method in future endeavors. 

Corresponding code, supplementary material, and a testable version of our model are available at: \url{https://github.com/foger3/ARC_DeepLearning}.

\begin{acknowledgments}
  Special thanks to Konrad Mikalauskas, Gustaw Opie{\l}ka, Luke Korthals, Leonhard Volz, Catherine Guazzone, Theresa Leidinger, Vincent Ott, Fridtjof Petersen, Milla Pihlajamäki, Arne John, Jan Failenschmid, Madlen Hoffstadt, Henrik Godmann, and Thomas Willems for their advice and comments. 
\end{acknowledgments}

%% Define the bibliography file to be used
\bibliography{bibliography}

\clearpage
\appendix{
\section{ConceptARC accuracy table}
\begin{table}[htb!]
  \caption{Accuracy on the Concept Groups introduced by ConceptARC across Rule Vector (RV) Approaches and Humans \cite{Moskvichev23}} 
  \label{tab:concept}
  \centering
  \begin{tabular}{cccc}
    \toprule
    Concept & Humans & Average RV & Similarity RV \\
    \midrule
    Above and Below & 0.90 & 0.00 & 0.00 \\
    Center & 0.94 & 0.07 & 0.03 \\
    Clean Up & 0.97 & 0.07 & 0.03 \\
    Complete Shape & 0.85 & 0.00 & 0.00 \\
    Copy & 0.94 & 0.03 & 0.00 \\
    Count & 0.88 & 0.53 & 0.60 \\
    Extend To Boundary & 0.93 & 0.00 & 0.00 \\
    Extract Objects & 0.86 & 0.07 & 0.07 \\
    Filled and Not Filled & 0.96 & 0.03 & 0.03 \\
    Horizontal and Vertical & 0.91 & 0.00 & 0.00 \\
    Inside and Outside & 0.91 & 0.10 & 0.03 \\
    Move To Boundary & 0.91 & 0.00 & 0.00 \\
    Order & 0.83 & 0.03 & 0.00  \\
    Same and Different & 0.88 & 0.07 & 0.07 \\
    Top and Bottom 2D & 0.95 & 0.00 & 0.07 \\
    Top and Bottom 3D & 0.93 & 0.00 & 0.00 \\
    \bottomrule
\end{tabular}
\end{table}
}

%% End of file
\end{document}